%% file: emnlp2020.tex
\documentclass[11pt,a4paper]{article}
\usepackage{emnlp2020}
\usepackage{times}
\usepackage{stfloats}
\usepackage{latexsym}

\usepackage{microtype}
\aclfinalcopy 

\newcommand*\samethanks[1][\value{footnote}]{\footnotemark[#1]}

\title{What Have We Achieved on Text Summarization? }

\author{
\textbf{Dandan Huang\textsuperscript{\rm 1,2}\thanks{Equal contribution.}\hspace{1.5mm},
Leyang Cui\textsuperscript{\rm 1,2,3}\samethanks\hspace{1.5mm},
Sen Yang\textsuperscript{\rm 1,2}\samethanks\hspace{1.5mm}},\\
\textbf{Guangsheng Bao\textsuperscript{\rm 1,2},
Kun Wang,
Jun Xie\textsuperscript{\rm 4},
Yue Zhang\textsuperscript{\rm 1,2}\thanks{Corresponding author.}\hspace{1.5mm}}\\
\textsuperscript{1} School of Engineering, Westlake University \\
\textsuperscript{2} Institute of Advanced Technology, Westlake Institute for Advanced Study\\
\textsuperscript{3} Zhejiang University,
\textsuperscript{4} Tencent SPPD\\

{\tt \{huangdandan, cuileyang, yangsen, baoguangsheng\}@westlake.edu.cn},\\
{\tt wongkhun@outlook.com}, 
{\tt stiffxie@tencent.com},
{\tt yue.zhang@wias.org.cn}
}

\date{}

\usepackage{times}
\usepackage{latexsym}
\usepackage{amsfonts,amssymb}
\usepackage{booktabs}
\usepackage{amsmath}
\usepackage{multirow}
\usepackage{tabularx}
\usepackage{graphicx}
\usepackage[lofdepth,lotdepth]{subfig}
\usepackage{bm}
\usepackage{tcolorbox}
\usepackage[marginal]{footmisc}
\usepackage{framed}
\usepackage{color}
\usepackage{soul}
\usepackage{colortbl}

\usepackage[dash,dot]{dashundergaps}
\usepackage{diagbox}

\begin{document}
\maketitle
\begin{abstract}
Deep learning has led to significant improvement in text summarization with various methods investigated and improved ROUGE scores reported over the years. However, gaps still exist between summaries produced by automatic summarizers and human professionals. Aiming to gain more understanding of summarization systems with respect to their strengths and limits on a fine-grained syntactic and semantic level, we consult the Multidimensional Quality Metric$\footnote{MQM is a framework for declaring and describing  human writing quality which stipulates a hierarchical listing of error types restricted to human writing and translation.}$ (MQM) and quantify 8 major sources of errors on 10 representative summarization models manually. Primarily, we find that 1) under similar settings, extractive summarizers are in general better than their abstractive counterparts thanks to strength in faithfulness and factual-consistency; 2) milestone techniques such as copy, coverage and hybrid extractive/abstractive methods do bring specific improvements but also demonstrate limitations; 3) pre-training techniques, and in particular sequence-to-sequence pre-training, are highly effective for improving text summarization, with BART giving the best results.
\end{abstract}

\section{Introduction}
Automatic text summarization has received constant research attention due to its practical importance. Existing methods can be categorized into extractive~\cite{dorr-etal-2003-hedge,textrank,aaai2017:summarunner} and abstractive~\cite{jing-2000-cut,rush-2015-neural,acl2017:pointer-generator}
methods, with the former directly selecting phrases and sentences from the original text as summaries, and the latter synthesizing an abridgment by using vocabulary words. Thanks to the resurgence of deep learning, neural architectures have been investigated for both extractive~\cite{neuralsum,xu-durrett-2019-neural} and abstractive~\cite{seq2seq-attention-summary,bart,structsum} summarization systems. 

Although improved ROUGE scores have been reported on standard benchmarks such as Gigaword~\cite{gigaword}, NYT~\cite{NewYorkTime} and CNN/DM~\cite{CNN-DailyMail} over the years, it is commonly accepted that the quality of machine-generated summaries still falls far behind human written ones. 
As a part of the reason, ROUGE has been shown insufficient as a precise indicator on summarization quality evaluation~\cite{liu-liu-2008-correlation,bohm-etal-2019-better}. In the research literature, human evaluation has been conducted as a complement~\cite{narayan-etal-2018-ranking}. However, human evaluation reports that accompany ROUGE scores are limited in scope and coverage. On a fine-grained level, it still remains uncertain what we have achieved overall and what fundamental changes each milestone technique has brought. 

We aim to address the above issues by quantifying the primary sources of errors over representative models. In particular, following MQM~\cite{MQM}, we design 8 metrics on the {\it Accuracy} and {\it Fluency} aspects. Models are analyzed by the overall error counts on a test set according to each metric, and therefore our evaluation can be more informative and objective compared with existing manual evaluation reports. We call this set of metrics {\bf PolyTope}. Using PolyTope, we manually evaluate 10 text summarizers including Lead-3, TextRank~\cite{textrank}, Sequence-to-sequence with Attention~\cite{rush-2015-neural}, SummaRuNNer~\cite{aaai2017:summarunner}, Point-Generator~\cite{acl2017:pointer-generator}, Point-Generator-with-Coverage~\cite{coveragemechanism,acl2017:pointer-generator}, Bottom-Up~\cite{bottom-up}, BertSumExt~\cite{emnlp2019:bertsum}, BertSumExtAbs~\cite{emnlp2019:bertsum} and BART~\cite{bart}, through which we compare neural structures with traditional preneural ones, and abstractive models with their extractive counterparts, discussing the effectiveness of frequently-used techniques in summarization systems. Empirically, we find that:
\begin{itemize}
    \item[1.] Preneural vs Neural: Traditional rule-based methods are still strong baselines given powerful neural architectures.
    \item[2.] Extractive vs Abstractive: Under similar settings, extractive approaches outperform abstractive models in general. The main shortcoming is {\it unnecessity} for extractive models, and {\it omission} and {\it intrinsic hallucination} for abstractive models.
    \item[3.] Milestone Techniques: Copy works effectively in reproducing details. It also reduces duplication on the word level but tends to cause redundancy to a certain degree. Coverage solves repetition errors by a large margin, but shows limits in faithful content generation. Hybrid extractive/abstractive models reflect the relative strengths and weaknesses of the two methods.
    \item[4.] Pre-training: Pre-training is highly effective for summarization, and even achieves a better content selection capability without copy and coverage mechanisms. Particularly, joint pre-training combining text understanding and generation gives the most salient advantage, with the BART model achieving by far the state-of-the-art results on both automatic and our human evaluations.
\end{itemize}

We release the test set, which includes 10 system outputs and their manually-labeled errors based on PolyTope, and a user-friendly evaluation toolkit to help future research both on evaluation methods and automatic summarization systems$\footnote{https://github.com/hddbang/PolyTope}$.

\section{Related Work}
\label{related work}
\paragraph{Extractive Summarization}
Early efforts based on statistical methods~\cite{NetoFK02,textrank} make use of expertise knowledge to manually design features or rules. Recent work based on neural architectures considers summarization as a word or sentence level classification problem and addresses it by calculating sentence representations~\cite{neuralsum,aaai2017:summarunner,xu-durrett-2019-neural}. Most recently, ~\citet{zhong-etal-2020-extractive} adopts document-level features to rerank extractive summaries.

\paragraph{Abstractive Summarization}
Jing and McKeown~\shortcite{jing-2000-cut} presented a cut-paste based abstractive summarizer, which edited and merged extracted snippets into coherent sentences. Rush  et  al.~\shortcite{rush-2015-neural} proposed a sequence-to-sequence architecture for abstractive summarization. Subsequently, Transformer was used and outperformed traditional abstractive summarizer by ROUGE scores ~\cite{contrastive}. Techniques such as AMR parsing~\cite{AMR}, copy~\cite{copying-mech}, coverage~\cite{coveragemechanism,acl2017:pointer-generator}, smoothing~\cite{smoothing} and pre-training~\cite{bart,emnlp2019:bertsum} were also examined to enhance summarization. Hybrid abstractive and extractive methods adopt a two-step approach including content selection and text generation~\cite{bottom-up,hybrid-model,DCA}, achieving higher performance than end-to-end models in ROUGE. 
\paragraph{Analysis of Summarization}
There has been much work analyzing summarization systems based on ROUGE. Lapata and Barzilay~\shortcite{lapata2005automatic} explored the fundamental aspect of ``coherence" in machine generated summaries. Zhang et al.~\shortcite{emnlp2018:abstractiveness} analyzed abstractive systems, while Kedzie et al.~\shortcite{emnlp2018/contentselection} and Zhong et al.~\shortcite{analysis_extractive_2} searched for effective architectures in extractive summarization. Kryscinski et al.~\shortcite{2019-evaluation} 
evaluated the overall quality of summarization in terms of redundancy, relevance and informativeness.
All the above rely on automatic evaluation metrics.
Our work is in line with these efforts in that we conduct a fine-grained evaluation on various aspects. 
Different from the above work, we use human evaluation instead of automatic evaluation. In fact, while yielding rich conclusions, the above analytical work has also exposed deficiencies of automatic toolkits. The quality of automatic evaluation is often criticized by the research community~\cite{DBLP:conf/emnlp/NovikovaDCR17,zopf-2018-estimating} for its insufficiency in neither permeating into the overall quality of generation-based texts~\cite{liu-liu-2008-correlation} nor correlating with human judgements~\cite{2019-evaluation}.

There has also been analysis work augmenting ROUGE with human evaluation \cite{narayan-etal-2018-ranking,emnlp2019:bertsum}. Such work reports coarse-grained human evaluation scores which typically consist of 2 to 3 aspects such as informativeness, fluency and succinctness. Recently, \newcite{maynez2020faithfulness} conducted a human evaluation of 5 neural abstractive models on 500 articles. Their main goal is to verify the faithfulness and factuality in abstractive models. In contrast, we evaluate both rule-based baselines and extractive/abstractive summarizers on 8 error metrics, among which faithfulness and factuality are included.

Our work is also related to research on human evaluation for summarization. To this end, Pyramid~\cite{pyramid} scores a summarizer based on its system output and multiple references. Annotators are requested to identify the smallest content units of semantic meaning, and then associate each unit with a weight by counting how many reference summaries contain this unit. The score of a summary is computed according to the number and weight of units. In addition to Pyramid, there are human evaluation metrics based on ranking \cite{narayan-etal-2018-ranking}, best-worst scaling~\cite{best-worst} and question answering~\cite{QAmethod}. The above methods assign one score to each summarization output. In contrast to these methods, our error-count based metrics are motivated by MQM for human writing, and are more fine-grained and informative. We show more empirical contrast between evaluation metrics in Figure~\ref{annotation example} in Section~\ref{Analysis of Evaluation Methods}. Most recently, ~\citet{stiennon2020learning} uses human evaluation as a reward for training automatic summarizers, reporting significant improvement compared with models trained using reference summaries. Their work also demonstrates the usefulness of human evaluation in text summarization.

\section{Models}
\label{models}
We re-implement and evaluate 10 representative and influential methods. Their publicly reported ROUGE F1 scores are illustrated in Table~\ref{tab:overallROUGE}.
\begin{table*}
    \centering
    \resizebox{0.96\textwidth}{!}{
    \begin{tabular}{c|cccc|cccccc}
         \hline
         \multirow{2}{*}{\diagbox{ROUGE}{Methods}}& \multicolumn{4}{c|}{Extractive Methods} & \multicolumn{6}{c}{Abstractive Methods} \\
         \cline{2-11}
         & \bf{Lead-3} & \bf{TextRank} & \bf{Summa}&\bf{BertExt} & \bf{S2S} & \bf{PG} & \bf{PG-Coverage} & \bf{Bottom-Up} & \bf{BertAbs} & \bf{BART}\\
         \hline
         ROUGE-1 & 39.20 & 40.20 & 39.60 & 43.25 & 31.33 & 36.44 & 39.53 & 41.22 & 42.13 & {\bf 44.16}\\
         ROUGE-2 & 15.70 & 17.56 & 16.20 & 20.24 & 11.81 & 15.66 & 17.28 & 18.68 & 19.60 & {\bf 21.28} \\
         ROUGE-L & 35.50 & 36.44 & 35.30 & 39.63 & 28.80 & 33.42 & 36.38 & 38.34 & 39.18 & {\bf 40.90} \\
         \hline
    \end{tabular}
    }
    \caption{ROUGE scores of 10 summarizers on CNN/DM Dataset (non-anonymous version). We get the score of Lead-3 and TextRank from~\citet{aaai2017:summarunner} and~\citet{zhou-etal-2018-neural}, respectively.}
    \label{tab:overallROUGE}
    \vspace{-10pt}
\end{table*}

\subsection{Extractive Methods}
\paragraph{Lead-3} Lead-3 is a commonly-used baseline, which simply selects the first three sentences as the summary. It is used as a standard baseline by most recent work~\cite{neuralsum,bottom-up}. Intuitively, the first three sentences of an article in news domain can likely be its abstract, so the results of Lead-3 can be a highly faithful approximation of human-written summary.
\paragraph{TextRank} TextRank~\cite{textrank} is an unsupervised key text units selection method based on graph-based ranking models~\cite{pagerank}. It defines ``recommendation'' by calculating co-similarity between sentences and yielding a weighted graph accordingly. Sentences with high weights are extracted as summaries. It is selected as a representative of statistical models.
\paragraph{SummaRuNNer}
SummaRuNNer~\cite{aaai2017:summarunner} is a representative neural extractive model which selects full sentences from the input as a summary. It first encodes the input with a hierarchical BiGRU, then scans input sentences from left to right. An accumulated summary representation is generated by a weighted sum of all previous selections, which is fed into a logistic classifier to make the final prediction on summary.
\paragraph{BertSumExt}
\label{sec:bertsumext}
BertSumExt~\cite{emnlp2019:bertsum} takes pre-trained BERT~\cite{bert} as the sentence encoder and an additional Transformer as the document encoder. A classifier on sentence representation is used for sentence selection. It takes advantages of knowledge from fine-tuned BERT for generating better summaries.
\subsection{Abstractive Methods}

\paragraph{Seq2Seq with Attention} The sequence-to-sequence model structure is first used for abstractive summarization by Rush et al.~\shortcite{rush-2015-neural}. To allow effective and free text generation rather than simple selection and rearrangement, a target-to-source attention module is adopted to capture the information from every encoder hidden state. We follow the implementation of See et al.~\shortcite{acl2017:pointer-generator}. 
\paragraph{Pointer-Generator} \citet{acl2017:pointer-generator} introduces the pointer network \cite{pointernetwork} to address the problem that seq2seq models tend to reproduce factual details inaccurately. The method can both generate words from the vocabulary via a generator, and copy content from the source via a pointer.
\paragraph{Pointer-Generator-with-Coverage} See et al.~\shortcite{acl2017:pointer-generator} use the coverage mechanism~\cite{coveragemechanism} to avoid repetition problems. This mechanism calculates a coverage vector as an extra input for the attention mechanism to strengthen attention to different locations. 
\paragraph{Bottom-Up} Gehrmann et al.~\shortcite{bottom-up}  propose a two-step approach, first selecting potential output words and then generating a summary based on pointer-generator network. Bottom-Up is selected as a representative of hybrid models which integrate extractive and abstractive methods.
\paragraph{BertSumExtAbs}
\label{sec:bertsumextabs}
BertSumExtAbs~\cite{emnlp2019:bertsum} adopts the same encoder as BertSumExt, and a 6-layer Transformer decoder with randomly initialized parameters. It is selected as a representative of neural abstractive models with pretrained contextualized sentence representation.
\paragraph{BART}
Instead of pre-training the encoder only, BART~\cite{bart} jointly pre-trains a seq2seq model combining a bidirectional encoder and an auto-regressive decoder. Further fine-tuned on summarization datasets, it achieves the current state-of-the-art result in terms of ROUGE scores.


\section{Evaluation Method}
We analyze system performance by using ROUGE \cite{rouge} for automatic scoring and PolyTope for human scoring. ROUGE has been adopted by most work on summarization. It is a recall-based metric calculating lexical overlap between system output and human summaries. Particularly, ROUGE-1 is based on unigram overlaps, ROUGE-2 on bigrams and ROUGE-L on longest common subsequences. 

\begin{figure}
\centering
\includegraphics[width=0.4\textwidth]{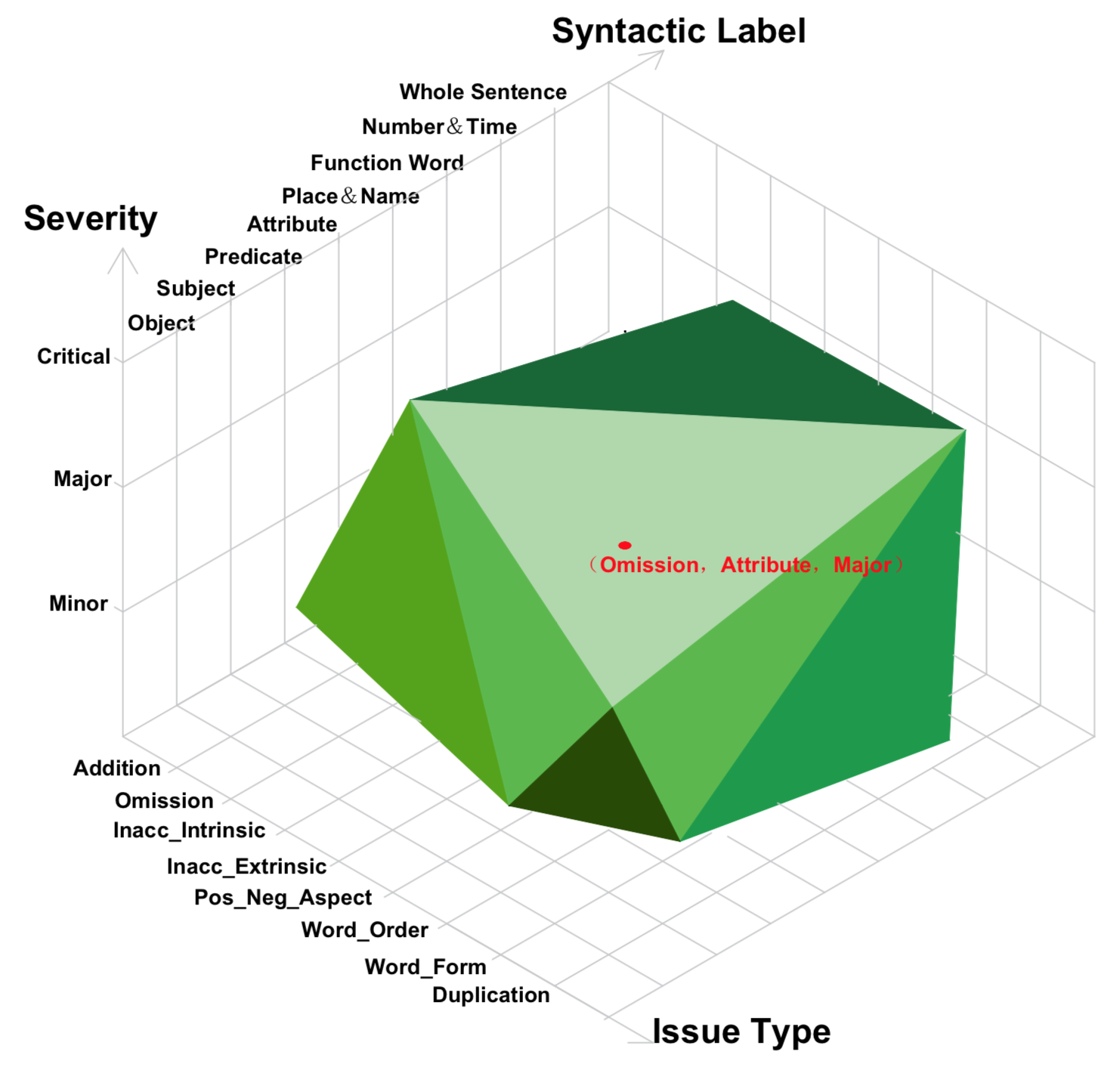}
\caption{PolyTope verdicts each error by three coordinates according to its syntactic and semantic roles.}
\label{fig:polytope}
\end{figure}

PolyTope is an error-oriented fine-grained human evaluation method based on Multidimensional Quality Metric (MQM)~\cite{MQM}. In particular, it consists of 8 issue types (Section~\ref{sec:issue_type}), 8 syntactic labels (Section~\ref{sec:syntactic_label}) and a set of severity rules (Section~\ref{severity}) to locate errors and to automatically calculate an overall score for the tested document. As illustrated in Figure~\ref{annotation example}, compared with ROUGE, PolyTope is more fine-grained in offering detailed and diagnostic aspects of overall quality. 

We develop an operating interface for annotation, which is shown in Appendix A.1. Particularly, a human annotator is presented the original text and an output summary in juxtaposition, and is asked to select segments that are deemed incorrect after reading. Upon a preliminary selection, he is asked to make a further selection among 8 issue types and 8 syntactic labels, respectively. An embedded severity score is then generated automatically for every incorrect segment, and the quality score is calculated for the annotated summary as:
\begin{equation*}
\text{Score} = (1 - \frac{\sum_{\alpha \in I}\alpha*{\text{Severity}_\alpha}}{\text{word}_{\text{count}}})*100,
\end{equation*}
where $I \in \{{\textsc{Minor}}{\textsc{,}} {\textsc{Major}}{\textsc{,}} {\textsc{Critical}}\}$,
indicating the error count for each severity. ${\text{Severity}}$ scores are deducted for errors of different severity, with the deduction ratio being set as 1:5:10 for \textsc{Minor}, \textsc{Major} and \textsc{Critical}, respectively. ${\text{word}_{\text{count}}}$ is the total number of words in samples. For a skilled annotator, it takes 2.5-4 minutes averagely to complete annotation of one sample, of which 2-3 minutes are used for extensive reading and 0.5-1 minutes for annotation. After PolyTope evaluation, 3-dimensional error points show the overall quality of the tested model (Figure~\ref{fig:polytope}). The inter-annotator agreement over 20 documents is 0.8621 in terms of Pearson correlation  coefficient, which shows that PolyTope can significantly reduce subjective bias of annotators. More human annotation details are illustrated in Appendix B.

\subsection{Issue Type}
\label{sec:issue_type}
Issue types of PolyTope can be categorized into {\it Accuracy} and {\it Fluency} issues, whose definitions can be traced to the MQM principle. {\it Accuracy}-related issues refer to the extent to which the content conveyed by the target summarization does not match or accurately reflect the source text. It comprises five sub-types:
\paragraph{Addition} Unnecessary and irrelevant snippets from the source are included in the summary.  
\paragraph{Omission} Key point is missing from the output.
\paragraph{Inaccuracy Intrinsic} Terms or concepts from the source are misrepresented and thus unfaithful. 
\paragraph{Inaccuracy Extrinsic} The summary has content not presented in the source and factually incorrect. 
\paragraph{Positive-Negative Aspect} The output summary represents positive statements whereas the source segment is negative, and vice versa.\\

{\it Fluency} issues refer to linguistic qualities of the text. Unlike {\it Accuracy}, {\it Fluency} is independent of the relationship between the source and the target. 
It comprises three sub-types:
\paragraph{Duplication} A word or longer portion of the text is repeated unnecessarily.
\paragraph{Word Form} Problems in the form of a word, including agreement, POS, tense-mood-aspect, etc.
\paragraph{Word Order} Problems in the order of syntactic constituents of a sentence.

Their examples are elaborated in Appendix A.2.

\begin{table*}\small
    \centering
    \resizebox{0.96\textwidth}{!}{
    \begin{tabular}{c|c|c|c|c|c|c|c|c|cc}
    
         \cline{1-10}
         \bf{Issue type}& \bf{Sub Issue Type} & \bf{Subject} & \textsc{\bf{Object}}&\textsc{\bf{Predicate}} & \bf{Number\&Time} & \bf{Place\&Name} & \bf{Attribute} & \bf{Function Word} & \textsc{\bf{Whole Sentence}}\\
         \hline
          \multirow{6}{*}{\bf{Accuracy}}& Addition & Critical & Critical & Critical & Major & Major & Major & Minor & Major \\
        & Omission & Critical & Critical & Critical & Critical & Major & Major & Minor & Critical \\
        & Inacc Intrinsic & Critical & Critical & Critical & Critical & Critical & Major & Minor & N/A \\
        & Inacc Extrinsic & Critical & Critical & Critical & Critical & Critical & Critical & Minor & N/A \\
        & Pos Neg Aspect & N/A & N/A & Critical & N/A & N/A & Critical & N/A & N/A\\ 
        \hline
        \multirow{3}{*}{\bf{Fluency}}
        & Word Order & N/A & N/A & Major & N/A & N/A & Major & Minor & N/A \\ 
        & Word Form & Minor & Minor & Minor & Minor & Minor & Minor & Minor & N/A \\
        & Duplication & Major & Major & Major & Major & Major & Major & Minor & Major \\
         \hline
     \end{tabular}
    }
    \caption{PolyTope for summarization diagnostics. This error matrix avoids subjectivity as human judgers only need to annotate issue types and syntactic labels of each mistake. Severity rules and scores is predefined and automatically calculated, without providing their own preference and scores.}
    \label{Matrix}
\end{table*}

\subsection{Syntactic Label}
\label{sec:syntactic_label}
Syntactic labels aim to locate errors, allowing tighter relevance between error issues and sentence constituents. According to ACE2005 (Automatic Content Extraction), we define 8 syntactic labels to distinguish sentence components, namely {\it Subject}, {\it Predicate}, {\it Object}, {\it Number\&Time}, {\it Place\&Name}, {\it Attribute}, {\it Function Word} and {\it Whole Sentence}. Their definitions are elaborated in Appendix A.3. 


\subsection{Severity}
\label{severity}
Severity is an indication of how severe a particular error is. It has three levels: \textsc{Minor, Major} and \textsc{Critical}, calculated by the evaluation tool automatically given the human decision on the error type and syntactic label. In practice, each cell in Table~\ref{Matrix} corresponds to a specific severity level. Issues with higher severity have more impact on perceived quality of the summary. 
\paragraph{Minor} Issues that do not impact usability or understandability of the content. For example, if grammar function word repeats itself, the redundant preposition is considered an error but does not render the text difficult to use or problematic.
\paragraph{Major} Issues that impact usability or understandability of the content but do not render it unusable. For example, an additional attribute may result in extra effort for the reader to understand the intended meaning, but does not make the content unfit for purpose. 
\paragraph{Critical} Issues that render the content unfit for use. For example, an omitted subject that changes the meaning of the text would be considered critical. If the error prevents the reader from using the content as intended or if it presents incorrect information that could result in harm to the user, it must be categorized as critical. In general, even a single critical error is likely to cause serious problems.

\section{Evaluating Model Performance}
\label{sec:analysis}
\begin{table*} \small
    \centering
    \resizebox{0.96\textwidth}{!}{
    \begin{tabular}{c|cccc|cccccc}
         \hline
         & \multicolumn{4}{c|}{Extractive Methods} & \multicolumn{6}{c}{Abstractive Methods} \\
         \hline
         & \bf{Lead-3} & \bf{TextRank} & \bf{Summa}&\bf{BertSumExt} & \bf{S2S} & \bf{PG} & \bf{PG-Coverage} & \bf{Bottom-Up} & \bf{BertSumExtABS} &\bf{BART}
         \\
         \hline
         ROUGE-1 & 41.63 & 33.81 & 41.11 & 42.69 & 31.87 & 38.89 & 39.90 & 41.19 & 41.87 & 43.28\\
         ROUGE-2 & 19.62 & 13.71 & 20.15 & 21.19 & 13.07 & 19.64 & 19.00 & 19.98 & 21.02 & 21.28\\
         ROUGE-L & 35.55 & 26.47 & 36.40 & 35.95 & 29.48 & 35.92 & 35.01 & 36.52 & 34.16 & 38.13\\
         ROUGE-1 Rank & \hl{{\bf\#4}}
         & \#9 & \#6 & \hl{\bf\#2} & \#10 & \#8 & \#7 & \#5 & \hl{{\bf\#3}} &\hl{{\bf\#1}} \\
         \hline
         Addition & 329 & 272 & 156 & 160 & 125 & 117 & 143 & 207 & 165 & 135\\
         Omission & 196 & 309 & 193 & 185 & 329 & 286 & 256 & 287 & 213 & 115\\
         Inacc\_Intrinsic & 0 & 0 & 0 & 0 & 304 & 14 & 16 & 68 & 7 & 2\\
         Inacc\_Extrinsic & 0 & 0 & 0 & 0 & 42 & 0 & 0 & 4 & 0 & 0\\
         Pos\_Neg\_Aspect & 0 & 0 & 0 & 0 & 3 & 0 & 0 & 3 & 0 & 0\\
         \hline
         Word\_Order & 0 & 0 & 0 & 0 & 0 & 0 & 0 & 0 & 0 &0\\
         Word\_Form & 0 & 0 & 0 & 0 & 1 & 0 & 0 & 0 & 1 &0\\
         Duplication & 17 & 12 & 36 & 9 & 139 & 68 & 11 & 6 & 3 & 2\\
         \hline
         Critical & 192 & 302 & 191 & 184 & 588 & 284 & 257 & 333 & 213 & 112\\
         Major & 350 & 289 & 194 & 170 & 317 & 193 & 161 & 210 & 172 & 140\\
         Minor & 0 & 2 & 0 & 0 & 38 & 8 & 8 & 32 & 4 & 2\\
         Errors / 1k Words & 55 & 61 & 39 & 37 & 160 & 70 & 56 & 84 & 48 & 30
         \\
         \hline
         PolyTope Score & 81.96 & 77.07 & 85.43 & 86.03 & 36.61 & 72.55 & 77.80 & 67.99 & 81.52 & 89.37\\
         PolyTope Rank & \hl{{\bf\#4}} & \#7 & \hl{{\bf\#3}} & \hl{{\bf\#2}} & \#10 & \#8 & \#6 & \#9 & \#5 & \hl{{\bf\#1}}\\
         \hline
    \end{tabular}
    }
    \caption{ROUGE and PolyTope results on 150 instances from CNN/DM dataset. ROUGE is the F1 score with stemming and stopwords not removed, giving the best agreement with human evaluation.}
    \label{tab:performance}
\end{table*}

We evaluate the aforementioned 10 models using the above two metrics, focusing on comparisons between pre-neural and neural methods, extractive and abstractive methods, and better understanding the effects of milestone techniques such as copy, coverage, pre-training and hybrid abstractive/extractive models.
We randomly sample 150 trials from the non-anonymized CNN/DM dataset~\cite{CNN-DailyMail}. When predicting summaries, we select three sentences as the summary for extractive models following the original papers, and let the algorithms self-stop for abstractive models, which also give three sentences as the decoding result in most cases. Table~\ref{tab:performance} presents the performances based on PolyTope and ROUGE. Cases supporting observations below are illustrated in Appendix C.

\subsection{Preneural vs Neural Models}
On ROUGE-1, Lead-3 ranks the 2nd among extractive models, and the 4th among all models. On PolyTope, it ranks the 3rd among extractive models and the 4th among all models. This shows that the simple method stands as a strong baseline even among neural methods. TextRank ranks the 9th and 7th among all methods on ROUGE and PolyTope, respectively, still competitive to some abstractive neural models. On the negative side, these two methods show the largest numbers of {\it Addition} errors, which demonstrates that unsupervised methods are relatively weaker in filtering out useless information compared to the supervised methods.

\subsection{Extractive vs Abstractive Summarization}
On ROUGE, there is no strong gap between extractive and abstractive methods, with BART and BertSumExt being the top  abstractive and extractive models, respectively. 
On PolyTope, as a representative of abstractive models, BART overwhelmingly outperforms the others (p $<$ 0.01 using t-test). However, excluding BART, extractive models take the following top three places. Under similar settings, extractive methods are better (p $<$ 0.01 using t-test) compared with abstractive counterparts (e.g. BertSumExt vs BertSumExtAbs, SummaRuNNer vs Point-Generator, Point-Generator-with-Coverage).

Extractive models tend to make only 3 types of errors, namely {\it Addition}, {\it Omission}, {\it Duplication}, while abstractive models make 4 to 7 types of errors. With respect to {\it Accuracy}, extractive methods are notably stronger in terms of {\it Inacc Intrinsic} and {\it Extrinsic}, which reflects that through directly copying snippets from the source, extractive methods are guaranteed to produce a summary with fair grammaticality, rationality and loyalty. However, extractive methods do not show stronger performances in {\it Addition} and {\it Omission}, which is because extracted sentences contain information not directly relevant to the main points. With regard to {\it Fluency}, two approaches are generally competitive with each other, showing that nowadays neural models are relatively effective in synthesizing coherent summaries.

\subsection{Extractive Methods}
\label{Extractive Methods}
We first compare neural methods BertSumExt and SummaRuNNer. BertSumExt gives better ROUGE-1/2 compared to SummaRuNNer, but the difference is not significant under ROUGE-L or PolyTope. Among detailed errors, BertSumExt demonstrates advantages only in {\it Duplication}, for the likely reason that the contextualized representations of the same phrases can be different by BERT encoding. 
It co-insides with previous findings~\cite{emnlp2018/contentselection} which demonstrate that more complicated architectures for producing sentence representations do not lead to better performance under the setting of extractive summarization. 
Given the fact that gold-standard extractive summaries are constructed according to ROUGE, the better ROUGE score of BertSumExt reflects the effectiveness of stronger representation on fitting training data. 

\begin{figure}
\centering
\subfloat[Extractive models. \label{fig:coverage}]{\includegraphics[width=0.44\textwidth]{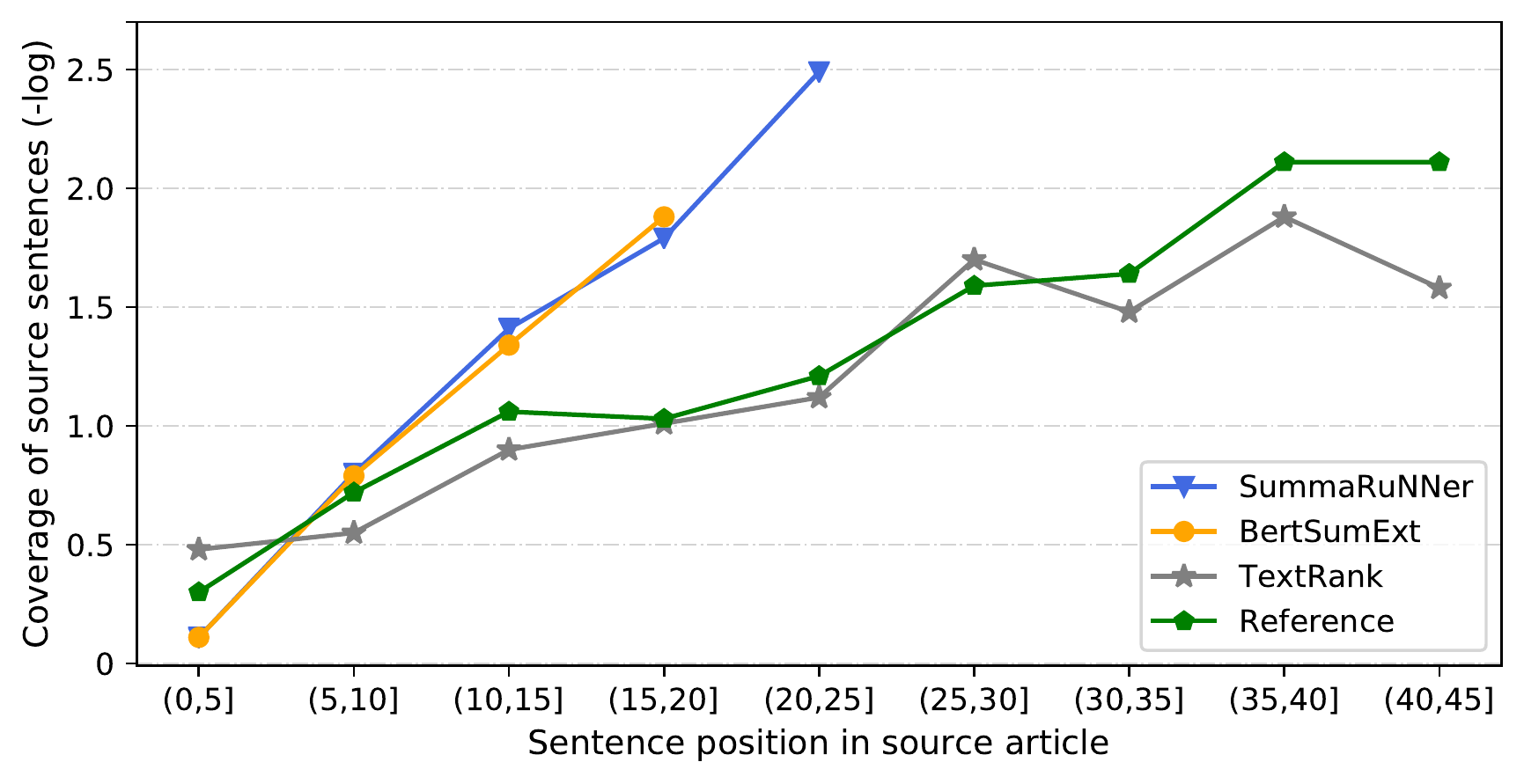}}
\vspace{-8pt}
\subfloat[Abstractive models. \label{coverage-abs}]{\includegraphics[width=0.44\textwidth]{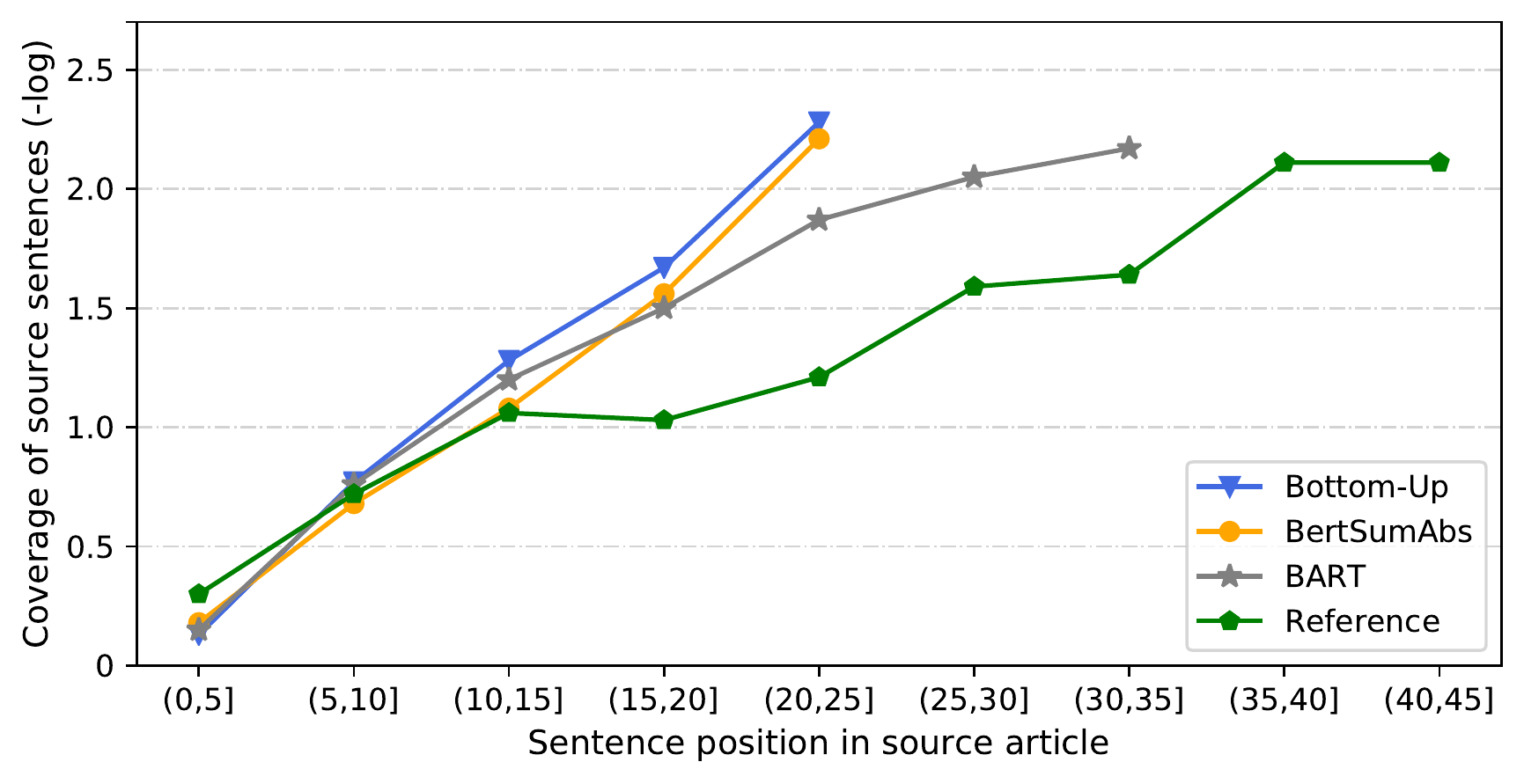}}
\caption{Distribution of source sentence used for content generation. X-axis: sentence position in source article. Y-axis: the negative log of coverage of sentence.}
\label{fig:length}
\vspace{-6pt}
\end{figure}

We then take statistical models into account. Figure~\ref{fig:coverage} shows the distribution of source sentences used for content generation by each method. There is a high proportion in the first five sentences and a smooth tail over all positions for reference summaries. In contrast, BertSumExt and SummaRuNNer extract sentences mostly from the beginning, thereby missing useful information towards the end. TextRank improves the coverage slightly as it is graph-based and does not depend on sequence information. But as lack of supervision, the model has a large number of {\it Addition} and {\it Omission}.

\subsection{Abstractive Methods}
\label{Abstractive Methods}
\paragraph{Copy} The na\"{i}ve seq2seq model suffers an {\it Inacc-Intrinsic} count of 304, the worst among all models compared. In contrast, the Point-Generator model reduces the error count to 14, demonstrating the effectiveness of the copy mechanism in faithfully reproducing details. Another interesting finding is that {\it Duplication} 
errors are also sharply reduced from 139 to 68, although the copy mechanism is not explicitly designed to address this problem. Further investigation shows that the reduced duplication patterns are mostly on the word level, while the effect on sentence-level duplication reduction is nearly zero. One likely reason is that the seq2seq decoder relies heavily on short-term history when deciding the next output word, without effective use of long-term dependencies. The Point-Generator model solves this problem by interpolating vocabulary level probability with copy probability, reducing reliance on previous outputs. On the negative side, the copy mechanism introduces {\it Addition} errors, because the auto-regressive point generator network tends to copy long sequences in entirety from the source, failing to interrupt copying at desirable length. This is also observed by Gehrmann et al.~\shortcite{bottom-up} and Balachandran et al.~\shortcite{structsum}.

\paragraph{Coverage} Coverage~\cite{coveragemechanism} is introduced to neural summarization systems to solve repetition issues. Compared with Point-Generator, Point-Generator-with-Coverage reduces {\it Duplication} errors from 68 to 11 and {\it Omission} errors from 286 to 256, proving that coverage is useful for better content selection. However, Point-Generator-with-Coverage yields more {\it Addition} and {\it Inacc\_Intrinsic} errors than Point-Generator. We further extract outputs of Point-Generator that do not have {\it Duplication} errors, finding that introducing the coverage mechanism reduces the average PolyTope scores from 77.54 to 74.07. It indicates that the coverage mechanism lacks inference capability and tends to generate summaries that incorrectly combine contents from the source into irrelevant information (see Figure10 and Figure11 in Appendix C as examples). This is likely because the coverage mechanism forces attention values from the decoder to the encoder to move monotonically to the right, and therefore can interfere with the original content selection process.

\begin{figure*}[t]\small
\centering
\includegraphics[width=\textwidth, height=300pt]{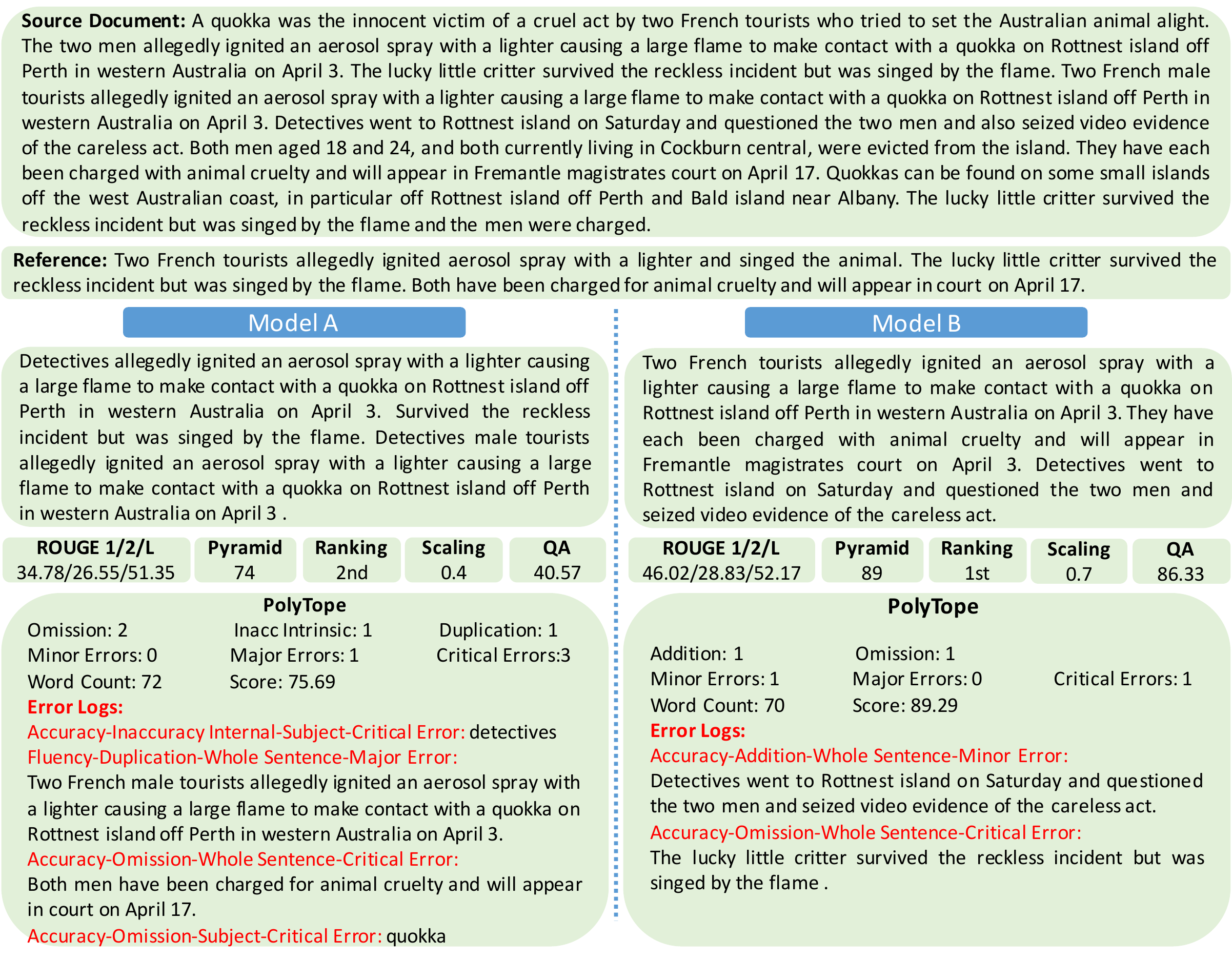}
\caption{A case study that compares various evaluation methods with each other.}
\label{annotation example}
\end{figure*}

\paragraph{Hybrid Abstractive/Extractive Model} Bottom-Up gives high ROUGE scores, but ranks the second worst on PolyTope. Compared with others, it suffers more from {\it Inaccuracy} errors. The inconsistency between ROUGE and PolyTope reflects the relative strengths and weaknesses of this method. On the positive side, it combines the advantages of extractive and abstractive models in selecting segments from the source and generating new contents in the summary, leading to a better recall. On the negative side, the abstractive generation model constrains copy attention only on the extracted snippets, thereby suffering from incomplete information sources for making inference and consequently lack of faithfulness and factual consistency.

\paragraph{Pre-training} 
Both BertSumExtAbs and BART outperform the non-pretraining abstractive models by a large margin. They differ from the other methods in two aspects, namely the Transformer architecture and contextualized knowledge. Since it has been shown that Transformer does not bring improved ROUGE compared with LSTM~\cite{bottom-up,analysis_extractive_2}, knowledge encoded by large-scale pre-training is likely the key for their better performance. Without the help of copy and coverage, BertSumExtAbs gives less number of {\it Inacc} and {\it Duplication} errors, and BART further gives the least number in almost all errors, showing the strength of pre-training technology. 

It is worth noting that BART ranks the 1st on both ROUGE and PolyTope among the 10 models. Different from BertSumExtAbs which pre-trains the encoder only, BART pre-trains the encoder and decoder jointly with seq2seq denoising auto-encoder tasks. It gives large improvements on {\it Addition}, {\it Omission} and {\it Inacc} errors, proving that unified pre-training for both understanding and generation is highly useful for content selection and combination. In particular, BART shows superior performance in handling the leading bias of CNN/DM dataset. Figure~\ref{coverage-abs} shows the distribution of source sentences used for content generation by the abstractive methods. As can be seen, abstractive models tend to neglect sentences in the middle and at the end of source documents (e.g., Bottom-Up, BertSumExtAbs), indicating that performance of abstractive summarizers is strongly affected by the leading bias of dataset. In contrast, BART can attend to sentences all around the whole document, slightly closer to the distribution of golden reference. Intuitively, this improvement might result from the document rotation transformation of BART pre-training, which shuffles the sentences on the encoder side for the same decoder. We leave the verification to future work, which requires re-training of BART without document rotation transformation. 



\begin{table}[t]  \centering
\begin{tabular}{l|c|cccc} 
\hline  
 & & R-1 & R-2 & R-L \\  
\hline
\multirow{3}{*}{\bf{Instance}}& PolyTope & 0.40 & 0.32 & 0.32\\
& Accuracy & 0.31 & 0.26 & 0.25 \\     
& Fluency & 0.07 & 0.41 & 0.01 \\    
\hline   
\bf{System} & PolyTope & 0.78 & 0.73 & 0.52 \\
\hline
 \end{tabular} 
\caption{Pearson correlation coefficients between ROUGE scores and human annotations from the perspective of instance and system level, respectively. }
  \label{tab:evaluation}
  \end{table}

\section{Analysis of Evaluation Methods}
\label{Analysis of Evaluation Methods}
The main goal of this paper is to investigate the differences between summarization systems, rather than to promote a human evaluation metric. Nonetheless, our dataset gives us a testbed to calculate the correlation between automatic and human evaluation methods. In this section, we report a contrast between ROUGE and PolyTope quantitatively, and between PolyTope and other human evaluation metrics qualitatively to demonstrate why we used PolyTope for our research goal.

First, research has shown that ROUGE is inconsistent with human evaluation for summary quality~\cite{liu-liu-2008-correlation,zopf-2018-estimating,2019-evaluation,maynez2020faithfulness}. We evaluate ROUGE using PolyTope from the perspective of both instance-level and system-level performances. On the instance level, the individual 1500 outputs from the 10 models are adopted to calculate the Pearson correlation coefficients between ROUGE and PolyTope. Additionally, we select test instances that only make {\it Accuracy} or {\it Fluency} errors to better understand the correlation between ROUGE and {\it Accuracy}/{\it Fluency} aspects. 
On the system level, the overall scores of each model are adopted to calculate the Pearson correlation coefficients between ROUGE and PolyTope. 

The results are summarized in Table~\ref{tab:evaluation}. For the instance-level comparison, we find a weak correlation between ROUGE and human judgement. In addition, with respect to {\it Accuracy} and {\it Fluency}, ROUGE can measure {\it Accuracy} to a certain extent, and ROUGE-2 is better than ROUGE-1/L in terms of evaluating {\it Fluency}. 
For the system-level comparison, the Pearson correlation coefficient is 0.78, 0.73, 0.52 for ROUGE-1, ROUGE-2, and ROUGE-L, respectively, much higher than 0.40, 0.32, 0.32 on the instance level. This confirms that ROUGE is useful for ranking systems after aggregation of samples but is relatively weak for assessing single summary quality, where the fine grained PolyTope could help~\cite{peyrard-etal-2017-learning}.

Second, Figure~\ref{annotation example} shows results of two models on one test document by ROUGE, Pyramid, ranking, scaling, QA and PolyTope evaluation metrics. As can be seen from the figure, PolyTope offers more fine-grained information in quality evaluation. ~\citet{sun-etal-2019-compare} warned that human evaluation prefers to give higher scores to longer and more informative summaries. Under the setting of PolyTope, there was relatively little influence from the sentence length. Taking BertSumExt and BertSumExtAbs models as examples, the Pearson correlation coefficients between length of their outputs and the corresponding scores is 0.25 and 0.27, respectively, suggesting that PolyTope is more objective and meaningful for current models that produce summaries without pre-specified length.

Finally, we also evaluate the reference summaries of our 150 test trials by means of PolyTope, obtaining a general score of 96.41, with 63 errors in the {\it Accuracy} aspect and 0 errors in the {\it Fluency} aspect. Gold summaries did not receive full marks in the PolyTope evaluation, mainly because of hallucinating content. For example, a news article describes an event as happening {\it ``on Wednesday"} in a summary although the original document has {\it``on April 1"}. The human summary requires external knowledge beyond the document and thus suffers penalization. Another common hallucination involves rhetorical but irrelevant sentences, e.g., {\it``Click here for more news"}. In addition, there are a few grammatical issues that affect the accuracy. For example, in {\it``Piglet was born in China with only two front legs has learned to walk.”}, there is a missing conjunction between two verb phrases.


\section{Conclusion}
We empirically compared 10 representative text summarizers using a fine-grained set of human evaluation metrics designed according to MQM for human writing, aiming to achieve a better understanding on neural text summarization systems and the effect of milestone techniques investigated recently. Our observations suggest that extractive summarizers generally outperform abstractive summarizers by human evaluation, and more details are also found about the unique advantages gained by copy, coverage, hybrid and especially pre-training technologies. The overall conclusions are largely in line with existing research, while we provide more details in an error diagnostics aspect.

\section*{Acknowledge}
We thank all anonymous reviewers for their constructive comments. This work is supported by NSFC 61976180 and a research grant from Tencent Inc.

\bibliography{emnlp2020}
\bibliographystyle{acl_natbib}
\input{appendix}

\end{document}

%% file: appendix.tex
\clearpage

\begin{appendix}

\section{Details on PolyTope}
\subsection{Annotation Toolkit}
\label{Appendixtoolkit}
\begin{figure*}[hb]
\centering
\includegraphics[width=\textwidth]{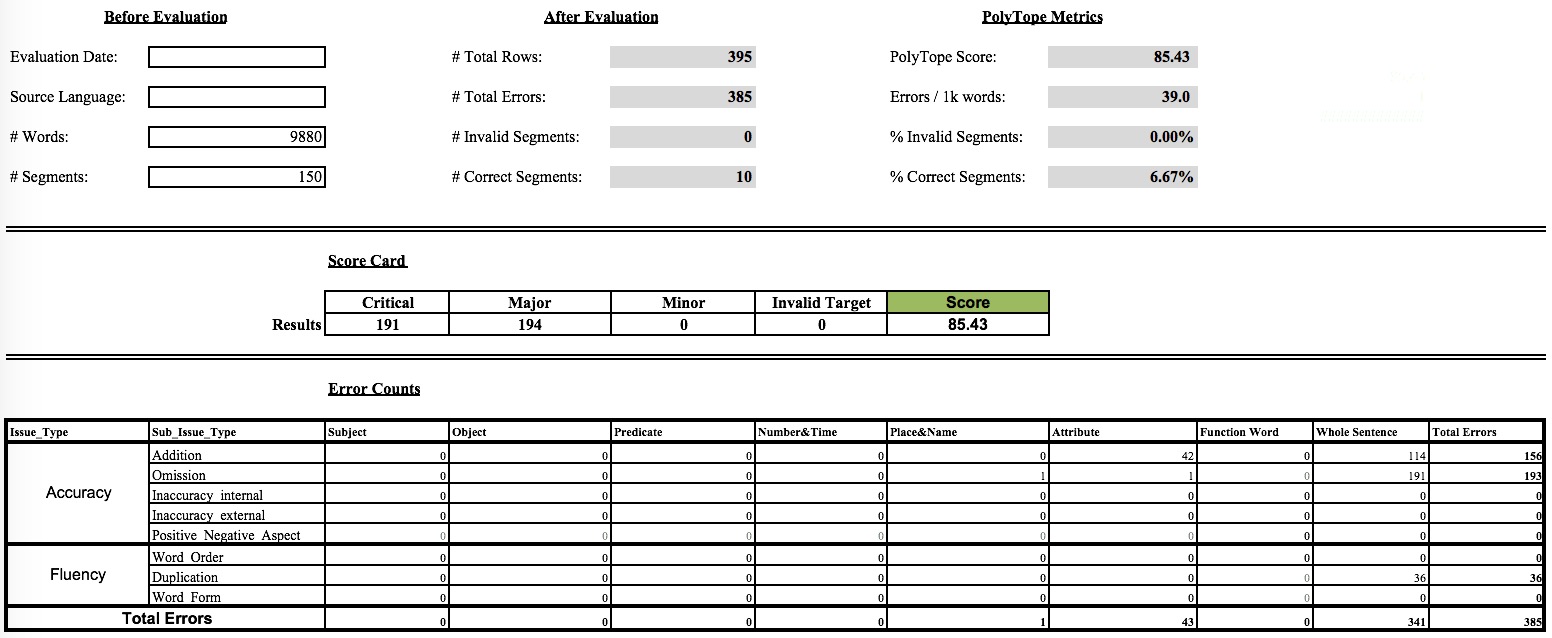}
\caption{Score Card}
\label{fig:scorecard}
\end{figure*}

\begin{figure*}[hb]
\centering
\includegraphics[width=\textwidth]{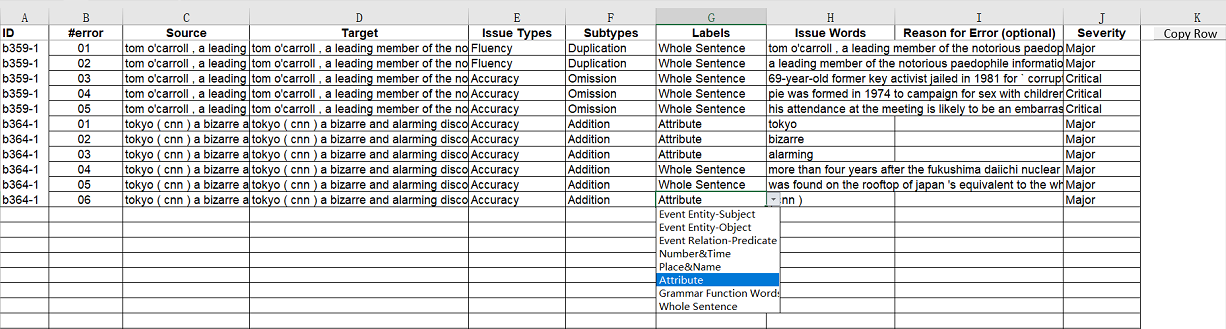}
\caption{Error Log}
\label{fig:errorlog}
\end{figure*}

\begin{figure*}[hb]
\centering
\includegraphics[width=\textwidth]{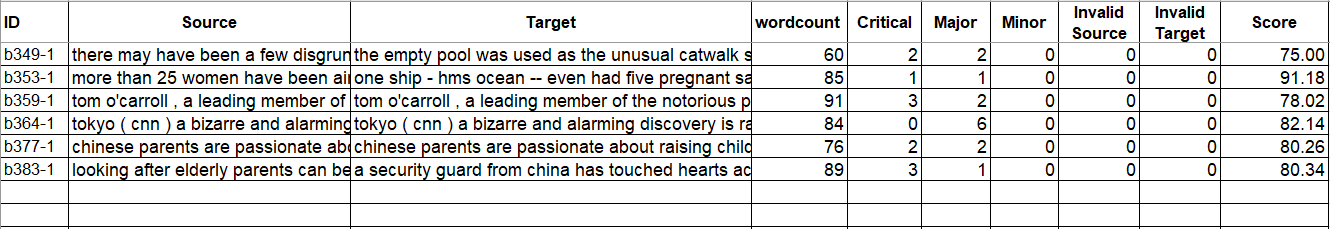}
\caption{Scores per Segment}
\label{fig:Scores per Segment}
\end{figure*}

We embed the evaluation rules in a Microsoft Excel workbook with Macros. The workbook contains 4 interrelated sheets, namely Score Card, Error Log, Scores per Segment, and Severity Matrix. 
\paragraph{Score Card} This sheet automatically calculates error numbers and scores (Figure~\ref{fig:scorecard}), and demonstrates the whole performance of the tested model. 

\paragraph{Error Log} This sheet is the annotation interface designed for annotators (Figure~\ref{fig:errorlog}). It is filled with source articles in column C and output summaries in column D in advance, and allows annotators to select segments that are deemed incorrect in column E to H.  Upon one selection, annotators are asked to make a selection among 8 issue types (column F) and 8 syntactic labels (column G). A severity is then generated automatically in column J, and a quality score is calculated automatically in the Scores per Segment sheet individually and in the Score Card sheet overall. 

\paragraph{Scores per Segment} This sheet calculates word count, error count and score for each tested sample (Figure~\ref{fig:Scores per Segment}). 

\paragraph{Severity Matrix} This sheet is the predefined severity matrix (Table~\ref{Matrix}) embedded in the Excel workbook by Macros.

\subsection{Issue Types}
\label{Issue Types}
We give examples on {\it Inaccuracy\_Intrinsic}, {\it Inaccuracy\_Extrinsic} and {\it Positive-Negative Aspect} as follows, as other errors are easy to understand.
\paragraph{Inaccuracy\_Intrinsic}
e.g., {\it``Pittsburgh Union Station is 10 kilometers from Exhibition Center and 3 kilometers from the University of Pittsburgh"} in the source but {\it``Pittsburgh Union Station is 3 kilometers from Exhibition Center"} in the output. 
\paragraph{Inaccuracy\_Extrinsic}
e.g., it is described as {\it``Pittsburgh Union Station, also known as Pittsburgh South Station"} in the output but {\it``Pittsburgh South Station"} is neither mentioned in the source text nor exists in the real world.

\paragraph{Positive-Negative Aspect}
e.g., {\it``push a button"} summarized as {\it``don't push a button"}, {\it``non-slip"} summarized as {\it``slip"}. This category applies only to actions and modifiers and refers to omitted or added negative particles (Figure~\ref{fig:233-2}).

\subsection{Syntactic Label}
\label{Appendixlabel}
\paragraph{Subject}	The event body that is being discussed.

\paragraph{Predicate} Acts and changes of state represented by verbs, verbal nouns, participles and gerunds. If a verb, verbal noun, participle or gerund acts as a modifier, it should be logged as ``Attribute".

\paragraph{Object} A noun or noun phrase that is affected by the action of a predicate.

\paragraph{Number\&Time} Number refers to digits, including cardinal and ordinal numerals, multiplicative and negative numbers, fractions and decimals, represented in numeric or word form. Time includes specific hours and minutes, part of the day (morning, evening, etc.), specific months and years, and words and phrases like tomorrow, in 3 days, during the next hours, etc.

\paragraph{Place\&Name} Place includes geographic name (e.g., {\it Europe}), administrative regions (e.g., {\it Texas}), specific addresses (e.g., {\it No 158, Fifth Ave}). Name includes names of real people, fictional characters, art, literature creations, companies, etc.

\paragraph{Attribute} Attribute refers to a syntax unit, either a word, phrase or clause, that modifies a noun.

\paragraph{Function Word} e.g., prepositions, auxiliary verbs, articles, determiners. 

\paragraph{Whole Sentence} A set of words that is complete in itself.

\section{Details on Human Evaluation }
Through a professional language service company, three candidates with a linguistic background and high level of English proficiency are employed for manual evaluation. They are all qualified language workers with satisfactory levels in reading, and pass the training and testing before being hired. They go through two pilot studies to have a better understanding of PolyTope framework and the nature of text summary. Documents used in the pilot studies are not used in the final annotation. During annotation, they are all naive to the model names, ROUGE scores, architectures and techniques of tested samples. Each of them is requested to annotate 50 instance, where one instance includes the original document and 10 model generated summaries. And then cross check. Overall, we have 1500 examples in total. Each successful completion includes annotation and quality inspection. In this manner, we try to not only ensure fairness but also assure the quality of human evaluation.

For TextRank, SummaRuNNer and BART, we implement the model strictly following their corresponding papers and achieve their reported ROUGE scores. Then models are used to produce summaries on the 150 test trails. For other models, we obtain summaries directly from their publicly available sources.

\section{Case Study}
Abstractive methods randomly splice fragments taken from the original text, leading to factual errors. See example of BertSumExtAbs (Figure~\ref{fig:5-6}), Pointer-Generator-with-Coverage (Figure~\ref{fig:exofsum}) and Bottom-Up (Figure~\ref{fig:42-4}).

Comparing Pointer-Generator and Pointer-Generator-with-Coverage, among outputs of Pointer-Generator that do not suffer from {\it Duplication} errors, introducing the coverage mechanism may interfere with the original content selection process and cause new problems. See examples in Figure~\ref{fig:28-3} and Figure~\ref{fig:15-3}.

The hybrid model gives high ROUGE scores overall, but does not necessarily combine strengths of extractive and abstractive methods. See example in Figure~\ref{fig:52-5}.

An example of {\it Positive-Negative Aspect} error is in Figure~\ref{fig:233-2}.

\section{Details on Layout Bias Calculation}
We compute the similarity score for each sentence in the output summary with each sentence in the source document by \textsc{BERTScore}~\cite{bertscore}, and illustrate a distribution of source sentence used for summary generation in Figure~\ref{fig:coverage} and Figure~\ref{coverage-abs}. In the news domain, neural summarization methods are typically biased towards selecting and generating summaries based on the leading paragraph of the document~\cite{emnlp2018/contentselection, analysis_extractive_2}. This stems from the structure of news articles, which present the most salient information of an article in the first few sentences and expand in the subsequent ones.

\section{Pearson Correlation}
ROUGE scores in Table~\ref{tab:evaluation} in the main text refer to the standard F1 scores. We also compute the Pearson correlation coefficients between ROUGE-P and PolyTope measurement, for the reason that ROUGE-P measures the precision which might be highly correlated to the proposed error-based PolyTope framework. We list the results in Table~\ref{P} for reference.

\begin{table}[h]
\centering
     \begin{tabular}{c|ccc}
          \hline
          & R-1 & R-2 & R-L \\
          \hline
          PolyTope & 0.15 & 0.22 & 0.06\\
          \hline
     \end{tabular}
     \caption{Pearson between ROUGE-P and PolyTope.}
     \label{P}
\end{table}

\section{Slips in Reference}
The CNN/DM dataset is a commonly used summarization dataset which contains news articles and associated highlights as summaries. We choose to focus on this dataset for the following reasons: First, both extractive and abstractive works report results on this benchmark dataset. Second, the gold summary in the dataset is the highlight sentence prefacing each article, which in most cases contains three sentences. This length is relatively closer to real world applications and more comprehensive for analysis than shorter summaries such as single-sentence summary. Hence, it provides us with a better benchmark to assess summarization models.

However, as a nature of the CNN/DM dataset, some reference summaries are of poor quality. Figure~\ref{fig:169-1} shows an exemplary document-summary pair whose summary contains grammatical errors. Figure~\ref{fig:5-1} shows an exemplary document-summary pair whose summary has noise like ``Click here for...". Figure~\ref{fig:62-1} shows an exemplary document-summary pair whose summary contains rhetorical sentences that interest readers but not crucial information to comprehend the document. In these cases, performance evaluation based on automatic evaluation is unreliable.

\clearpage
\onecolumn

\begin{figure*}[t]
    \centering
    \setlength{\fboxrule}{1pt}
    \setlength{\fboxsep}{0.5cm}
    \fbox{\parbox{\textwidth}{
        \textbf{Source Document}
        It used to be as much a part of a Sunday routine as eating a roast dinner or reading the papers. But new figures show that the art of washing your own car appears to be dying out in Britain with a third of men admitting they have never picked up a bucket or chamois leather to clean their own motor. The study also reveals that three-quarters of women never wash their own car with drivers more likely to take it to a car wash on a local forecourt. A new study has revealed that a third of men have never picked up a bucket or a chamois leather to wash their own car. The survey of 1,100 adults by vehicle leasing firm OSV, found that 31 percent of men have never washed their own car, with only \hl{12 percent} of those that do saying they \hl{do it regularly}. Meanwhile only \hl{5 percent} of those surveyed said that had ever \hl{asked their children to wash the car}, as a way for them to earn extra pocket money. Factors behind the decline vary from shops now opening on a Sunday and more live football on TV, meaning more people put off the chore at the weekend. The rise of hand car washes has been blamed for a decrease in drivers washing their own cars. Other reasons given for not washing cars include the rise of hand car washes and the fact that 4x4 and off-road vehicles are too large to clean. Brits are more likely to wash their own car during the holidays as the recent Easter break saw an estimated 15 million cars cleaned by their owners. However, one in four said they only did it as an excuse to get out of the house. Cars are most likely to be washed by their owners every week in Birmingham, Chelmsford and Sheffield and least likely in Oxford, Bath and Edinburgh. Audis, Toyotas and Bmws are most likely to be washed by their owners and Range Rovers, Peugeots and Mercedes least often. OSV spokesman Andrew Kirkley said: ``A lot of us remember seeing our streets full of men cleaning their cars every Sunday. It was a national tradition. But let's face it, we don't have as much time any more, the cars are bigger and even the kids don't seem to want to earn a bit of extra money doing it for us either." On top of that, there is now plenty of choice out there if you want to put the car through an automated machine or get a gang of enthusiastic guys to do it fairly cheaply.\\[0\baselineskip]
        \textbf{Reference}
        31 percent of British men say they have never washed their own car. Only 12 percent that have cleaned their own car say they do it regularly.

        \textbf{Model Output}
        Three-quarters of women never wash their own car with drivers more likely to take it to a car wash on a forecourt. Survey of 1,100 adults by vehicle leasing firm OSV found 31\% of men have never washed their own vehicle. Only \hl{12\%} of those surveyed said that had ever \hl{asked their children to wash the car}, as a way for them to earn extra pocket money.
    }}
    \caption{Factual errors made by the BertSumExtAbs model.}
    \label{fig:5-6}
\end{figure*}

\begin{figure*}[t]
    \centering
    \setlength{\fboxrule}{1pt}
    \setlength{\fboxsep}{0.5cm}
    \fbox{\parbox{\textwidth}{
        \textbf{Source Document}
            She might be approaching her 89th birthday but nothing was going to stop the queen from enjoying a ride in the spring sunshine this morning. The monarch, 88, was spotted trotting through Windsor great park aboard her faithful black fell pony, Carltonlima Emma. Joined by a groom on another of her fell ponies, the queen cut a relaxed figure as she enjoyed her ride but, as is her wont, eschewed a helmet in favor of one of her silk scarves. Enjoying the sunshine: the queen enjoys a ride on her fell pony Carltonlima Emma. \hl{The queen, who has never worn riding helmets, has been encouraged to wear the safety hats in the past but is reportedly reluctant to wear one because of her hair}. Speaking in an interview last year, her racing trainer Ian balding recalled the moment he asked why the monarch never wears a riding hat. The queen is said to have replied: ``I never have and you don’t have to have your hair done like I do." Her majesty is famous for her love of horses and first found herself in the saddle at the age of four after being presented with a Shetland pony, named Peggy, aged four. Since then, the royal stables have been home to a succession of steeds, among them Betsy, a black farm-bred horse who was her mount of choice in the 50's, and surprise, a grey gelding whom the queen famously galloped down the course at ascot in 1961. Equine enthusiast: her majesty adores the ponies and breeds them at Hampton court. No helmet: the queen never wears a riding helmet, preferring instead to ride in a silk headscarf. Cutting back: she has ridden less in recent years as a result of a niggling knee injury. Long term love: the queen has ridden all her life and continues to breed several breeds of horse and pony. Recent years have seen her cut down on the amount of time she spends in the saddle - the result of a niggling knee injury that also forced her to give up presiding over trooping the color on horseback. Nevertheless, the queen remains an enthusiastic equestrienne and, according to sources, is a familiar sight at her Windsor stables. She is also said to take a keen interest in all her horses and ponies, some of whom are now ridden by her grandchildren, notably Prince Edward's children, Lady Louise and James, Viscount Severn. Along with her thoroughbred race horses, the queen also breeds fell ponies and has a stud specialising in highland ponies at balmoral. First love: the queen's first pony was a tiny Shetland named Peggy who was given to her at the age of four. Familiar sight: the queen riding her much-loved horse Burmese during trooping the color. Seal of approval: a fell pony foal similar to those being bred by the queen at Hampton court. One of the oldest equine breeds on the planet, fell ponies like the queen's mount Carltonlima Emma, have roamed the moors of Cumbria since the Neolithic period. Docile, hardy and thick-set, the majority of fell ponies are black, grey or bay and measure between 13 and 14hh. Highly prized by the Carvetii, the iron age tribe who occupied the region more than two millennia ago, the ponies later caught the eye of the invading Romans and were used as trade goods all over the empire. The Romans also helped develop the breed into the relatively large animal it is today. Originally thought to have stood around 12hh tall (similar to Dartmoor and Exmoor ponies), by the end of the roman period, the average fell was more than a hand higher. Another invader to fall in love with the fell pony was the vikings, who used the animals as pack ponies; a use continued by the Normans. By the 13th century, the fell's usefulness as a pack animal was well-established and the pony played an important role in British trade until the end of the 18th century. Following the industrial revolution, fells were used to transport iron ore and coal from mine to town, as well as underground when the height of the shaft allowed. Although previously used in trotting races, the fell pony really came into its as a riding horse in the 1950s, when its gentle nature and pretty looks made it the pony of choice for families. Today, the pony remains a popular choice for riders of all ages, among them the queen who breeds fell ponies at her Hampton court stud. Source: the fell pony society.
      \newline
      \textbf{Reference}
           The queen was spotted enjoying a ride in Windsor great park today. Rode her favorite fell pony, named Carltonlima Emma. Left hard hats at home and opted for one of her favorite scarves instead.
      \newline
      \textbf{Model Output}
          The monarch, 88, was spotted trotting through Windsor great park aboard her faithful black fell pony, Carltonlima Emma. \hl{The queen, who has never worn riding helmets, has been encouraged to wear one because of her hair.}
        }}
    \caption{Factual errors made by the Point-Generator-with-Coverage model.}
    \label{fig:exofsum}
\end{figure*}

\begin{figure*}[t]
    \centering
    \setlength{\fboxrule}{1pt}
    \setlength{\fboxsep}{0.5cm}
    \fbox{\parbox{\textwidth}{
        \textbf{Source Document}
            An argument between two Brisbane neighbors over noise levels has ended in tragedy with one man dead and another charged with manslaughter. Leon Yeaman, 55, was allegedly killed by a single punch in the head from 28-year-old shift worker Phillip Pama, 28, during a row in front of their north Brisbane homes. Mr.Yeaman was reportedly in his yard on muller road making a lot of noise at around 9 pm Thursday night. \hl{Phillip Pama} covered his face as he left Brisbane magistrate courts where he was \hl{granted conditional bail on Friday}. Pama, a married father of three, works a 3 am shift at a nearby warehouse and was sleeping at the time Mr.Yeaman was making a commotion. Pama's defence lawyer Bruce peters said Mr.Yeaman was known for getting drunk, playing loud music and yelling at ``the world at large". Pama then confronted Mr.Yeaman and it was at that point the situation escalated. Mr.Peters said his client grabbed his neighbor by the hand and hit out ``blindly". ``He was being pushed at the same time he was throwing the punch," Mr.Peters revealed during a bail application for Pama. ``he didn't see it connect but he felt it connect." The father of three allegedly killed his 55-year-old neighbor with one punch after he refused to turn his music down. Paramedics were unable to revive Mr.Yeaman who was pronounced dead at the scene at around 11 pm on Thursday night. Neighbors described Mr.Yeaman as king and generous while his friend John Fisher told ABC radio he was a schizophrenic pensioner who enjoyed playing loud music. Pama shielded his face from reporters outside Brisbane magistrate courts when he was granted bail on Friday. The Samoan man was ordered to surrender his passport, aherre to a curfew from 6 pm to 2 am and will have to report to the local police station twice a week. Pama, who has no history of violence, was reportedly ``beside himself" with grief. The case is due to return to court on May 18.
      \newline
      \textbf{Reference}
           Two men got in an argument about noise levels on Thursday night. A scuffle broke out and Phillip Pama struck Leon Yeaman once in the head. Mr.Yeaman could not be revived and was announced dead at the scene. Pama has no history of violence and is reportedly ``beside himself". His defence lawyer said he hit out ``blindly" and didn't see his punch connect. Pama was granted conditional bail and is set to return to court May 18.
      \newline
      \textbf{Model Output}
          Leon Yeaman, 55, was allegedly killed by a single punch in the head from 28-year-old shift worker Phillip Pama, 28, during a row in front of their north Brisbane homes. \hl{Mr.Yeaman was granted conditional bail on Friday. He was ordered to return to court on May 18}.
        }}
    \caption{Factual errors made by the Bottom-Up model.}
    \label{fig:42-4}
\end{figure*}

\begin{figure*}[t]
    \centering
    \setlength{\fboxrule}{1pt}
    \setlength{\fboxsep}{0.5cm}
    \fbox{\parbox{\textwidth}{
        \textbf{Source Document}
            An obsessive arsenal fan made his girlfriend sit a test quizzing her about his favourite football club. Marking her answers with red pen, she scored an impressive 43.5 out of 50, to which he commented, ``Well done Saskia! As a result, I will not be dumping you! (lucky you)". Saskia posted snaps of the test to her twitter account, @saskiaisabxlla, commenting ``boyfriend made me take an @arsenal exam!" with emoticons showing she didn't know whether to laugh or cry. We don't blame her. Saskia's football fan boyfriend made her take an arsenal test and didn't dump her thanks to her A grade. The test took the form of a school exam, with multiple choice questions, tables to complete, and 'in your own words...' section. There were even ``helpful" instructions including ``read every question at least twice" and ``check your answers if you have time at the end". Oh, and ``calculators may be used". Saskia scored an impressive 87 percent, but according to her unnamed boyfriend's feedback she could have performed better on q17, needs to learn all player positions and should understand that tottenham are scum. She scored some decent points for her more creative answers. Saskia clearly knows how to keep her boyfriend happy, as she humoured his question of ``why are arsenal the best team to ever grace english football?" with the answer of ``they've won many fa cups" and ``Thierry Henry is a legend". However, when she got an answer wrong, her examiner wrote the correct answer in capitals and drew an angry face. Saskia struggled to name all 11 players of the starting squad and got confused on the player positions. He was impressed with the end result, though, grading her an ``A" and rewarding her by not ending their relationship. Comments generally hailed the man as a ``legend" and fellow football fans were impressed with Saskia's knowledge. Some twitter users hailed the unnamed boyfriend who tested Saskia on her arsenal knowledge as a legend. scoring 87\%, football fans were impressed with Saskia, claiming her as ``proper girlfriend material".
      \newline
      \textbf{Reference}
            Arsenal fan's girlfriend avoids getting dumped after passing football test. Saskia got an ``A", but apparently needs to work on her player positions. The bemused woman posted pictures of boyfriend's test on twitter.
      \newline
      \textbf{Model Output (PG)}
            Saskia posted snaps of the test to her twitter account, @saskiaisabxlla. She posted snaps of the test to her twitter account, commenting ``Well done Saskia! As a result, I will not be dumping you! (lucky you)".
      \newline
      \textbf{Model Output (PG Coverage)}
            Arsenal fan made his girlfriend sit a test quizzing her about his favourite football club. She posted snaps of the test to her twitter account, @saskiaisabxlla, commenting ``boyfriend made me take an @arsenal exam!" with emoticons showing she didn't know whether to laugh or cry. \hl{Oh, and ``calculators may be used"}.
        }}
    \caption{The Pointer-Generator-with-Coverage model tends to make {\it Addition} errors when Pointer-Generator does not have repetitions.}
    \label{fig:28-3}
\end{figure*}

\begin{figure*}[t]
    \centering
    \setlength{\fboxrule}{1pt}
    \setlength{\fboxsep}{0.5cm}
    \fbox{\parbox{\textwidth}{
        \textbf{Source Document}
            A teenager set herself on fire after allegedly being raped by five men from her village in India – and her family reportedly told her to keep quiet about the attack. The 14-year-old is now fighting for her life in Delhi’s Safdarjung hospital with 70 percent burns. She was allegedly gang-raped on Sunday when she went outside her house in Kosi Kalan, in Uttar Pradesh's Mathura district, to relieve herself. A teenager set herself on fire after allegedly being raped by five men from her village in India. She is now recovering in Delhi’s Safdarjung hospital. She was allegedly gang-raped on Sunday when she went outside her house in Kosi Kalan, in Uttar Pradesh's Mathura district, to relieve herself. On Tuesday \hl {she set herself on fire using kerosene, according to NDTV}, to the shock of her brother, who doused her with water. He told the broadcaster: ``when I woke up, I saw her in flames ... I poured water on her to put out the flames." \hl {The girl was screaming for help, according to a neighbour}. The accused men have been arrested. There is a heightened sensitivity to the issue of sexual assault in India at the moment after officials last month banned India's daughter, a documentary about the gang rape and murder of an Indian student in Delhi. Recent attacks in India have resulted in street protests with many calling for more protection for women. Officials said the documentary would cause further disorder if it was shown, following a number of protests and incidents of vigilante justice in the country. The documentary explained the brutal rape and murder of 23-year-old student Jyoti Singh, who was attacked on a bus when she returned home from the cinema. One of the six men convicted of the attack, bus driver Mukesh Singh, was interviewed in prison and told researchers that had Jyoti not fought back she would not have been killed. Her death led to protests throughout India and outraged the world. Last month an angry mob was seen on video footage beating a man to death in the street who was accused of raping and murdering an 11-year-old girl. Video footage has emerged of the brutal prolonged attack on the 18-year-old, which was watched by a jeering 1,000-strong crowd in Nagaland in eastern India. Ibo Cha was said to have been beaten for an hour before he died of his injuries. The footage was shot in September last year after the girl's body was found in woodland, enraging locals. But it only came to light after earlier this year alleged rapist Syed Sarif Khan was kidnapped from prison and dragged through the streets of the same area. He was then stripped naked and beaten to death. He was accused but not convicted of raping a 19-year-old female student multiple times. Later Nagaland government said he was innocent. Jyoti Singh Pandey, a physiotherapy student, was gang raped as she travelled on a bus. The 23-year-old suffered in hospital for 13 days from her injuries before she died. Vinay Sharma, 20, Akshay Thakur, 28, Pawan Gupta, 19, and Mukesh Singh, 26, were all sentenced to death for her rape. Ram Singh, co-accused and widely considered the leader of the group, was found dead in his cell. A minor also found guilty was sentenced to three years in a reformatory institution. Her death sparked angry protests in India and internationally about misogyny in the country. The attention forced judges to prioritise the case and the lawyer's association in Saket reportedly refused to defend the perpetrators. \\[0\baselineskip]
      \textbf{Reference}
          She was allegedly attacked after leaving her house to relieve herself. The attack is said to have taken place in India's Uttar Pradesh region. Victim suffered 70 percent burns after dousing herself in kerosene. Her brother apparently saw her covered in flames and threw water on her. \\[0\baselineskip]
      \textbf{Model Output (PG)}
           The 14-year-old is now fighting for her life in Delhi’s Safdarjung hospital. She was allegedly gang-raped on Sunday when she went outside her house. She was allegedly gang-raped on Sunday when she went outside her house. She set herself on fire using kerosene, according to \hl{NDTV}. \\[0\baselineskip]
      \textbf{Model Output (PG Coverage)}
           {Teenager set herself on fire after allegedly being raped by five men. She was allegedly gang-raped on Sunday when she went outside her house in Kosi Kalan, Uttar Pradesh's Mathura district, to relieve herself. On Tuesday she set herself on fire using kerosene, according to a \hl{neighbor}}. 
    }}
    \caption{The Pointer-Generator-with-Coverage model tends to incorrectly combine information from the document, thus leading to {\it Inacc\_Intrinsic} errors. }
    \label{fig:15-3}
\end{figure*}

\begin{figure*}[t]
    \centering
    \setlength{\fboxrule}{1pt}
    \setlength{\fboxsep}{0.5cm}
    \fbox{\parbox{\textwidth}{
        \textbf{Source Document}
            A trend we are just starting to get our heads around is the wide leg trouser. Be it denim, cropped, printed or striped, the wide leg trouser is at the forefront of ss15 trends. There's something effortless about a wide leg trouser that really appeals. And if like us you are growing tired of the skinny jean and want to try out a new look this could be your answer. Louise redknapp says that a wide-leg trouser can come as a welcome relief from the skinny jean. Skinny jeans have held court for quite a few years now and while they will never go out of style the wide leg will give you an alternative look. It's not the first time this look has made a comeback since the seventies. Lou tried the out the trend seven years ago with a stella mccartney flared jean - luckily she kept hold of them! The wide leg can be intimidating, especially if you're on the shorter side. But the key is all in the shoe. Huge amounts of fabric is hard to look flattering but if you keep your heels high and your top half simple you can pull it off with confidence. A high heeled shoe with an exposed top of foot will elongate the leg, a must when wearing a cropped voluminous trouser. Louise and Emma say to avoid being drowned by too much fabric try a high heeled shoe. Louise recommends pairing a wide-leg with a simple knit to keep the outfit in proportion if going for a longer length go high-waisted and very long to give extra leg length. Proportion is everything, team with a simple knit, shirt or tee to keep the look current and cool, not costume. Denim is the easiest way to nail the trend - Gucci revisited the seventies with a collection filled with flared jeans in vintage washes. The stylish pair agree that denim is the easiest way to achieve the trend. The wide-leg trouser made an appearance on several catwalks including on the runway of Gucci ss15 who showcased a denim take on the trend, for a high street take try the topshop miller jean (right). Celine showcased colourful billowing bottoms on their ss15 runway topshop meanwhile have been quick to follow suit with their version - the miller - a 1970s-style wide leg flare. Celine and Stella Mccartney kept it classy with a dressier-style pant, perfect for an evening look. Try new online label finery London for a similar style with a lower price tag. A wide billowing trouser silhouette will update your wardrobe this season and the bonus is it's one spring trend you don't need nice weather for. To shop our looks and for more wide-leg trouser inspiration head to astylealbum.com. Poppy Delevingne channeled the trend at the aw15 chloe show in Paris.
        \newline
        \textbf{Reference}
           Louise Redknapp and stylist Emma Thatcher try the wide-leg trouser. They say it makes a good replacement for ever-popular skinny jeans. 
      \newline
      \textbf{Model Output (Bottom up)}
      Louise redknapp says trouser can come as the wide leg trouser jean. Lou jeans will not go out of the leg trouser. The trouser is a wide leg trouser.
       \newline
      \textbf{Model Output (\textsc{Summa})}
          Louise Redknapp says that a wide-leg trouser can come as a welcome relief from the skinny jean. Louise and Emma say to avoid being drowned by too much fabric try a high heeled shoe. Poppy Delevingne channeled the trend at the aw15 chloe show in Paris.
      \newline
      \textbf{Model Output (\textsc{BertSumExtABS})}
          The wide leg trouser is at the forefront of ss15 trends. It's not the first time this look has made a comeback since the seventies. Louise and Emma say to avoid being drowned by too much fabric try a high heeled shoe.
        }}
    \caption{The Bottom-Up works bad but other models work well.}
    \label{fig:52-5}
\end{figure*}

\begin{figure*}[t]
    \centering
    \setlength{\fboxrule}{1pt}
    \setlength{\fboxsep}{0.5cm}
    \fbox{\parbox{\textwidth}{
        \textbf{Source Document}
        (CNN) Bobbi Kristina Brown, the daughter of Bobby Brown and Whitney Houston, has ``global and irreversible brain damage", according to her grandmother. Though the 22-year-old is \hl{no longer in a medically induced coma}, she remains unresponsive, Cissy Houston said in a statement Monday after visiting her granddaughter. ``meeting with the doctors and understanding that she can live in this condition for a lifetime truly saddens me," Houston said. ``we can only trust in god for a miracle at this time.'' Houston's statement matched that from a source with knowledge of brown's condition, who told CNN on Monday that she remained in the same neurological state she has been in for nearly three months. She does not respond to visitors or familiar voices, and her eyes do not follow a person around the room, the source told CNN. She also has a tracheostomy in her throat, the source said. The reports come two days after Brown's father, Bobby Brown, said his daughter's condition had improved. ``i can say today, Bobbi is awake. She's watching me," Brown told the audience at Dallas' Verizon Theatre. The audience cheered. In a statement Monday, an attorney for the Brown family said that Bobbi Kristina Brown's condition has improved but that the kind of life she will lead remains to be seen. ``doctors have indicated that she will have a long life,'' attorney Christopher Brown said. ``however, Bobbi Kristina is presently embarking on a rehabilitation process, and the quality of her life will not be known for years to come.'' Who's who in the Bobbi Kristina Brown case? Bobby Brown was in an ``emotional state'' on stage when he made the remarks about his daughter being awake, according to the statement. ``she has made it out of ICU, opened her eyes and started a rehabilitation that will be long and hard,'' said Bobby Brown's wife, Alicia Etheredge Brown.
         \newline
         \textbf{Model Output}
         \hl{Bobbi Kristina Brown is in a medically induced coma}, her grandmother says. Bobbi Kristina Brown's condition has improved but that the kind of life will be seen. Brown's mother says Bobbi Kristina Brown's condition has improved.
                }}
    \caption{Example of positive-negtive errors.}
    \label{fig:233-2}
\end{figure*}

\begin{figure*}[t]
    \centering
    \setlength{\fboxrule}{1pt}
    \setlength{\fboxsep}{0.5cm}
    \fbox{\parbox{\textwidth}{
        \textbf{Source Document}
        This little piggy has become an Internet sensation after learning how to walk on just two feet. The piglet was born with its back legs missing and has mastered the art of balancing on its front trotters. The heartwarming clip featured on the people's daily online shows the young animal taking unsteady steps, wobbling about with its snout to the ground, foraging for food. Heartwarming: the two-legged pig has become an Internet sensation because of a cute video showing it learning to walk. There are several occasions it looked close to tipping over but remarkably the piglet always manages to regain its balance. It happily mixes with the rest of the litter, which have all been born with a full complement of legs. The piglet belongs to a farmer, Ms. Duan, from Qionglai city in Sichuan province, south western China. Since its birth, villagers have flocked to take a peek at the curious animal. Tricky: the piglet has to use extraordinary balance to stand on its front trotters while all its siblings were born with all four legs. Athletic: the piglet, nicknamed ``super pig" can balance on its two front trotters and has captured the hearts of thousands since a video went online. Siblings: one of the other piglets looks to copy the two-legged creature as it takes a quick rest from trotting about. Ms. Duan said: ``When the piglets were feeding I noticed something strange with one of them. When I picked it up I could see it had no hind legs." It is not the first animal with two legs instead of four to win plaudits in China. A rabbit at Liuzhou zoo became a star attraction after learning to walk on its front paws. Many tourists in China's Guangxi province have been taking a detour to see the furry star hop on its front paws. Looking for grub: the piglet balances on two legs while keeping its snout to the ground in the hunt for food.
        \newline
        \textbf{Reference}
        Piglet \hl{was} born in China with only two front legs \hl{has} learned to walk. Villagers have flocked to see the two-legged animal in Qionglai. A video released online \hl{and} has captured the hearts of thousands.
        }}
    \caption{Reference contains grammatical errors.}
    \label{fig:169-1}
\end{figure*}

\begin{figure*}[t]
    \centering
    \setlength{\fboxrule}{1pt}
    \setlength{\fboxsep}{0.5cm}
    \fbox{\parbox{\textwidth}{
        \textbf{Source Document}
         Conor Mcgregor is gearing up for a featherweight title challenge against Jose Aldo on July 11 and has unveiled a fearsome new tattoo ahead of the encounter. The 26-year-old Irishman has already got a picture of a gorilla eating a heart inked upon his chest and ``the notorious" has now revealed a tiger's face tattooed onto his stomach. ``If you see the tiger, it's too late. You're food." Mcgregor wrote next to the Instagram post of his new artwork. Conor Mcgregor reveals his new tattoo of a tiger on his stomach to his Instagram followers. Mcgregor (left) poses in the shop with a fan shortly after having his tattoo on his stomach done. Mcgregor is challenging Jose Aldo for his featherweight champion title in Las Vegas on July 11. Aldo makes the eighth defence of his belt against the Irish fighter in Las Vegas, but Mcgregor claimed last week that the man he is challenging lacks the same motivation as him. ``He doesn't want to be near me. He doesn't want this the way I want it." Mcgregor said. ``He can't hide the fact he doesn't want the belt in his presence." Conor Mcgregor grabbed Aldo's (left) belt when they took their promotional tour to Dublin. Mcgregor claims he has greater motivation to win the title than Aldo has to defend it for the eighth time. At the end of march, the duo were undergoing a promotional tour in Dublin when Mcgregor grabbed the belt from Aldo and raised it in front of 5,000 home supporters. The pair have a fractious relationship as it is, with a little under three months away until Mcgregor has a chance to legitimately hold the belt before his supporters.
     \newline
     \textbf{Reference}
         Conor Mcgregor shared a picture of his new tiger tattoo on his stomach. The 26-year-old Mcgregor is set to challenge Jose Aldo on July 11. Mcgregor grabbed Aldo's featherweight champion belt in Dublin. \hl{Click here for all the latest UFC news}.
     }}
    \caption{Noise data in reference.}
    \label{fig:5-1}
\end{figure*}

\begin{figure*}[t]
    \centering
    \setlength{\fboxrule}{1pt}
    \setlength{\fboxsep}{0.5cm}
    \fbox{\parbox{\textwidth}{
        \textbf{Source Document}
           Manchester United manager Louis Van Gaal says he has been dreaming of beating rivals City in Sunday's Derby at old Trafford but will have to do so without Robin Van Persie. Van Persie said he was fit to feature in the game against City on Sunday after ankle trouble but Van Gaal has ruled him out. ``Most of the players are fit but being fit for me is different." Van Gaal said. Robin Van Persie will not return to the fray for Manchester United against Manchester City on Sunday. Louis Van Gaal explained at his press conference on Friday that Van Persie is not yet fit enough to play. United players train in the sunshine ahead of their game against local rivals City on Sunday. ``Robin is not fit enough to play." With city rocking after defeat at crystal palace on Monday, Van Gaal and his players have an opportunity to finish as high as second or third and avoid a champions league qualifying fixture later in the summer. The United boss wants his side to move towards that target by taking all three points against City. Van Gaal said: ``I dream of it. Every player should dream of the victory. Of course I want to win because it's a big step up the table also. If we win then third place is available." Van Gaal watches on as he prepares his side for his first Manchester Derby at Old Trafford. Manager Van Gaal oversees the training while captain Wayne Rooney runs with the ball. Goal keeper David De Gea, Winger Angel Di Maria and Radamel Falcao were in training action. Rooney leads the way in training as he runs through some cones in the sunshine on Friday in Manchester. Di Maria, Juan Mata, Falcao, Marcos Rojo and Ander Herrera prepare to take on City. ``A few months ago, nobody would have thought about that, apart from me. If we win then the position in the table is good as we would then almost certainly be qualified for the top four. And if we are third it's better than the goal we set in preseason." Van Gaal acknowledged that City will provide a stern challenge to his team and played down the idea he will be motivated by revenge, after losing the away fixture earlier in the campaign. ``We have lost 1-0, that is my history, the last game," Van Gaal said. ``I say always in such games, always, you have to control your aggression." We did not do that at that time. So I hope we have learned from that moment. I can not say because we lost that game that we have to win this game. ``In my opinion, you can lose to Man City two times. That is possible." The united boss also ruled out a potential return for Luke Shaw. Like Van Persie, Van Gaal does not yet consider him fit enough to feature. ``I don’t think Shaw is fit enough to play," Van Gaal said. ``but I can not say they are not fit, but in my opinion they are not fit to play. That is a different question." Luke Shaw is another player yet to be at the standard of match fitness required by Van Gaal. Rooney, Ashley Young, Goalkeeper Anders Lindegaard and Michael Carrick have a breather. Falcao and Antonio Valencia look in high spirits as they prepare for the Derby. Herrera will be hoping to continue the fine form that has seen him become one of United's key players recently. Van Gaal admitted he is looking forward to sampling the Derby atmosphere at Old Trafford, as he takes on City with home advantage for the first time since arriving at the club. ``When I see the fans of Man UTD, they are supporting us in a marvellous way, I think," Van Gaal said. ``after matches, I give our fans a big compliment, not because I have to, because then I wouldn’t say it. I say it because I feel it."
        \newline
        \textbf{Reference}
        Manchester United face Manchester City in the premier league on Sunday. Robin Van Persie said he was fit for united after nearly two months out. But Louis Van Gaal has since revealed he will be without the striker. \hl{Van Persie declares himself fit, but do Manchester United need him?} Click here for all the latest Manchester United news.
        }}
    \caption{Reference contains rhetorical sentences that interest readers.}
    \label{fig:62-1}
\end{figure*}


\end{appendix}